\documentclass[11pt]{article}
\usepackage{url}
\usepackage{acl2016}
\usepackage{amsmath}
\usepackage{mathtools}
\usepackage{times}
\usepackage{latexsym}
\usepackage{tipa}
\usepackage{float}
\usepackage{booktabs}
\usepackage{graphicx,subcaption}
\usepackage{xcolor}
\usepackage{linguex}
\usepackage{tikz}
\usepackage{array}
\usepackage{comment}
\usetikzlibrary{calc,shapes,arrows,positioning,backgrounds}
\aclfinalcopy
\newcommand\blank{\underline{\hspace{0.5cm}}}
\newcommand{\argmax}{\operatornamewithlimits{argmax}}

\title{Issues in evaluating semantic spaces using word analogies}

\newif\ifanonymous
\anonymousfalse
\ifanonymous
\author{Author 1\\
	    XYZ Company\\
	    111 Anywhere Street\\
	    Mytown, NY 10000, USA\\
	    {\tt author1@xyz.org}
	  \And
	Author 2\\
  	ABC University\\
  	900 Main Street\\
  	Ourcity, PQ, Canada A1A 1T2\\
  {\tt author2@abc.ca}}
\else
\author{Tal Linzen\\
        LSCP \& IJN\\
	    \'Ecole Normale Sup\'erieure\\
        PSL Research University\\
	    \texttt{tal.linzen@ens.fr}
    }
\fi

\date{}

\begin{document}
\maketitle
\begin{abstract}

The offset method for solving word analogies has become a standard evaluation
tool for vector-space semantic models: it is considered desirable for a space
to represent semantic relations as consistent vector offsets. We show that the
method's reliance on cosine similarity conflates offset consistency with
largely irrelevant neighborhood structure, and propose simple baselines that
should be used to improve the utility of the method in vector space evaluation.

\end{abstract}

\section{Introduction}

\begin{figure}
    \centering
    \scalebox{0.8}{
    \begin{tikzpicture}[node distance=10mm and 5mm, framed]
        \node (a) {debug};
        \node [above right=of a.east] (aprime) {debugging};
        \draw [thick, red, ->] (a.north) to (aprime.south);
        \node [below right=5mm and 1cm of a] (b) {scream};
        \node [above right=of b.east, blue] (bprime) {screaming?};
        \draw [thick, red, ->] (b.north) to (bprime.south);
    \end{tikzpicture}
    }
    \caption{\label{fig:original}Using the vector offset method to solve the 
    analogy task \cite{mikolov2013linguistic}.}
\end{figure}
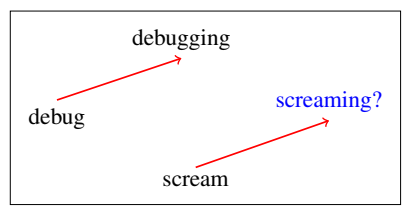

Vector space models of semantics (VSMs) represent words as points in
a high-dimensional space \cite{turney2010frequency}. There is considerable
interest in evaluating VSMs without needing to embed them in a complete NLP
system.  One such intrinsic evaluation strategy that has gained in popularity
in recent years uses the offset approach to solving word analogy problems
\cite{levy2014linguistic,mikolov2013linguistic,mikolov2013efficient,turney2012domain}.
This method assesses whether a linguistic relation --- for example, between the
base and gerund form of a verb (\textit{debug} and \textit{debugging}) --- is
consistently encoded as a particular linear offset in the space. If that is the
case, estimating the offset using one pair of words related in a particular way
should enable us to go back and forth between other pairs of words that are
related in the same way, e.g., \textit{scream} and \textit{screaming} in the
base-to-gerund case (Figure  \ref{fig:original}).

Since VSMs are typically continuous spaces, adding the offset between
\textit{debug} and \textit{debugging} to \textit{scream} is unlikely to land us
exactly on any particular word. The solution to the analogy
problem is therefore taken to be the word closest in cosine similarity
to the landing point. Formally, if the analogy is given by

\begin{equation}
    a : a^* :: b : \blank
\end{equation}

where in our example $a$ is \textit{debug}, $a^*$ is \textit{debugging} and
$b$ is \textit{scream}, then the proposed answer to the analogy problem is

\begin{equation}\label{eq:vanilla}
    x^* = \argmax\limits_{x'} \cos(x', a^* - a + b)
\end{equation}

where
\begin{equation}
    \cos(v, w) = \frac{v \cdot w}{\lVert v\rVert \lVert w \rVert}
\end{equation}

\begin{figure}
    \centering
    \includegraphics[width=0.95\linewidth]{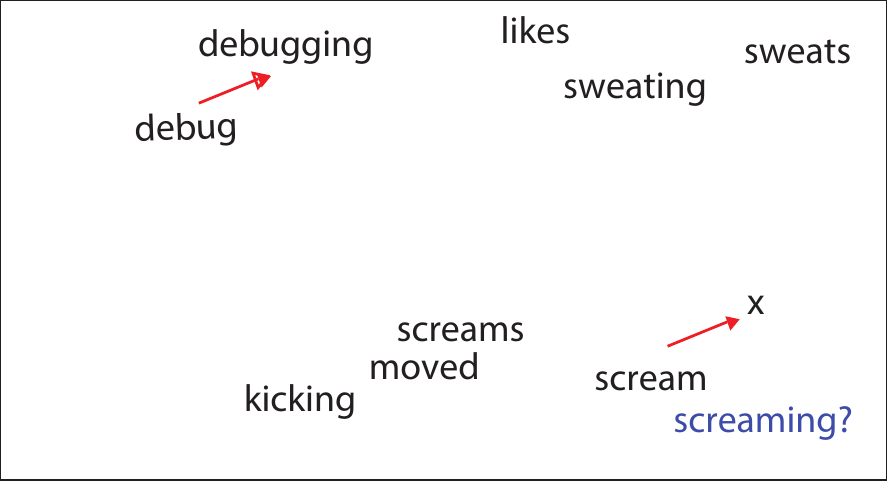}
    \caption{\label{fig:neighborhood_density}When $a^* - a$ is small and $b$ and $b^*$ are close, the expected answer may be returned even when the offsets are inconsistent (here \textit{screaming} is closest to $x$).}
\end{figure}

The central role of cosine similarity in this method raises the concern that
the method does not only evaluate the consistency of the offsets $a^* - a$ and
$b^* - b$ but also the neighborhood structure of $x = a^* - a + b$.  For instance, if
$a^*$ and $a$ are very similar to each other (as \textit{scream} and
\textit{screaming} are likely to be) the nearest word to $x$ may
simply be the nearest neighbor of $b$. If in a given set of analogies the
nearest neighbor of $b$ tends to be $b^*$, then, the method may give the
correct answer regardless of the consistency of the offsets 
(Figure \ref{fig:neighborhood_density}).

\begin{figure}
    \centering
    \scalebox{0.8}{
    \begin{tikzpicture}[node distance=10mm and 5mm, framed]
        \node (a) {debug};
        \node [above right=of a.east] (aprime) {debugging};
        \draw [thick, red, ->] (aprime.south) to (a.north);
        \node [below right=5mm and 1cm of a, blue] (b) {scream?};
        \node [above right=of b.east] (bprime) {screaming};
        \draw [thick, red, ->] (bprime.south) to (b.north);
    \end{tikzpicture}
    }
    \caption{\label{fig:reversed}Reversing the direction of the task.}
\end{figure}
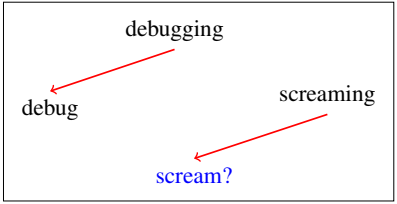

In this note we assess to what extent the performance of the offset method
provides evidence for offset consistency despite its potentially problematic
reliance on cosine similarity. We use two methods. First, we propose
new baselines that perform the task without using the offset $a^* - a$ and
argue that the performance of the offset method should be compared to those
baselines. Second, we measure how the performance of the method is affected by
reversing the direction of each analogy problem (Figure \ref{fig:reversed}). If
the method truly measures offset consistency, this reversal should not affect
its accuracy. 

\section{Analogy functions}

We experiment with the following functions.  In all of the methods, every word
in the vocabulary can serve as a guess, except when $a$, $a^*$ or $b$ are
explicitly excluded as noted below. Since the size of the vocabulary is
typically very large, chance performance, or the probability of a random word
in the vocabulary being the correct guess, is extremely low.

\paragraph{\textsc{Vanilla}:} This function implements the offset method
literally (Equation \ref{eq:vanilla}).  

\paragraph{\textsc{Add}:} The $x^*$ obtained from Equation \ref{eq:vanilla} is
often trivial (typically equal to $b$). In practice, most studies exclude
$a$, $a^*$ and $b$ from consideration:

\begin{equation}\label{eq:add}
    x^* = \argmax\limits_{x' \not\in \{a, a^*, b\}} \cos(x', a^* - a + b)
\end{equation}

\paragraph{\textsc{Only-b}:}

This method ignores both $a$ and $a^*$ and simply returns the nearest
neighbor of $b$:

\begin{equation}
    x^* = \argmax\limits_{x' \not\in \{a, a^*, b\}} \cos(x', b)
\end{equation}

As shown in Figure \ref{fig:neighborhood_density}, this baseline is likely to
give a correct answer in cases where $a^* - a$ is small and $b^*$ happens to be
the nearest neighbor of $b$.

\paragraph{\textsc{Ignore-a}:}

This baseline ignores $a$ and returns the word that is most similar
to both $a^*$ and $b$:

\begin{equation}
    x^* = \argmax\limits_{x' \not\in \{a, a^*, b\}} \cos(x', a^* + b)
\end{equation}

A correct answer using this method indicates that $b^*$ is closest to a point
$y$ that lies mid-way between $a^*$ and $b$ (i.e. that maximizes the similarity
to both words). 

\paragraph{\textsc{Add-opposite}:}

This function takes the logic behind the \textsc{Only-b} baseline a step
further -- if the neighborhood of $b$ is sufficiently sparse, we will get the
correct answer even if we go in the \textit{opposite} direction from the offset $a^*
- a$:

\begin{equation}
    x^* = \argmax\limits_{x' \not\in \{a, a^*, b\}} \cos(x', -(a^* - a) + b)
\end{equation}

\paragraph{\textsc{Multiply}:}

\newcite{levy2014linguistic} show that Equation \ref{eq:vanilla} is equivalent
to adding and subtracting cosine similarities, and propose replacing it with
multiplication and division of similarities:

\begin{equation}
    x^* = \argmax\limits_{x' \not\in \{a, a^*, b\}} \frac{\cos(x', a^*)\cos(x', b)}{\cos(x', a)}
\end{equation}

\paragraph{\textsc{Reverse (Add):}}

This is simply \textsc{Add} applied to the reverse analogy problem: if the
original problem is \textit{debug : debugging :: scream : \blank}, the reverse
problem is \textit{debugging : debug :: screaming : \blank}.  A substantial
difference in accuracy between the two directions in a particular type of
analogy problem (e.g., base-to-gerund compared to gerund-to-base) would indicate that the neighborhoods of
one of the word categories (e.g., gerund) tend to be sparser than the
neighborhoods of the other type (e.g., base).

\paragraph{\textsc{Reverse (Only-b):}} This baseline is equivalent to
\textsc{Only-b}, but applied to the reverse problem: it returns $b^*$, in the
notation of the original analogy problem.

\section{Experimental setup}

\paragraph{Analogy problems:}

\begin{table} 
\scalebox{0.8}{ 
    \begin{tabular}{l>{\it}l>{\it}lr} 
    \toprule
    ~ & $a$ & $a^*$ & $n$ \\
    \midrule
    Common capitals: & athens & greece & 506\\
    All capitals: & abuja & nigeria & 4524 \\
    US cities: & chicago & illinois & 2467 \\
    Currencies: & algeria & dinar & 866 \\
    Nationalities: & albania & albanian & 1599 \\ 
    Gender: & boy & girl & 506 \\
    Plurals: & banana & bananas & 1332 \\ 
    Base to gerund: & code & coding & 1056 \\ 
    Gerund to past: & dancing & danced & 1560 \\ 
    Base to third person: & decrease & decreases & 870 \\
    Adj. to adverb: & amazing & amazingly & 992 \\ 
    Adj. to comparative: & bad & worse & 1332 \\ 
    Adj. to superlative: & bad & worst & 1122 \\ 
    Adj. un- prefixation: & acceptable & unacceptable & 812 \\
    \bottomrule 
    \end{tabular}
}
\caption{\label{table:analogies}The analogy categories of
\newcite{mikolov2013efficient} and the number of problems per
category.} \end{table} 

We use the analogy dataset proposed by \newcite{mikolov2013efficient}. This
dataset, which has become a standard VSM evaluation set
\cite{baroni2014dont,faruqui2015retrofitting,schnabel2015evaluation,zhai2016intrinsic},
contains 14 categories; see Table \ref{table:analogies} for a full list.
A number of these categories, sometimes referred to as ``syntactic'', test
whether the structure of the space captures simple morphological relations,
such as the relation between the base and gerund form of a verb
(\textit{scream} : \textit{screaming}).  Others evaluate the knowledge that the
space encodes about the world, e.g., the relation between a country and its
currency (\textit{latvia} : \textit{lats}). A final category that doesn't
fit neatly into either of those groups is the relation between masculine and
feminine versions of the same concept (\textit{groom} : \textit{bride}).
We follow \newcite{levy2014linguistic} in calculating separate accuracy
measures for each category.

\paragraph{Semantic spaces:}

In addition to comparing the performance of the analogy functions within
a single VSM, we seek to understand to what extent this performance can differ
across VSMs. To this end, we selected three VSMs out of the set of spaces
evaluated by \newcite{linzen2016quantificational}. All three spaces were
produced by the skip-gram with negative sampling algorithm implemented in
word2vec \cite{mikolov2013distributed}, and were trained on the concatenation
of ukWaC \cite{baroni2009wacky} and a 2013 dump of the English Wikipedia.

The spaces, which we refer to as $s_2$, $s_5$ and $s_{10}$, differed only in
their context window parameters. In $s_2$, the window consisted of two words on
either side of the focus word. In $s_5$ it included five words on either side
of the focus word, and was ``dynamic'' -- that is, it was expanded if any of
the context words were excluded for low or high frequency (for details, see
\newcite{levy2015improving}). Finally, the context in $s_{10}$ was a dynamic
window of ten words on either side.  All other hyperparameters were set to
standard values.

\section{Results}

\paragraph{Baselines:}

\begin{figure}
    \centering
        \includegraphics[width=3in]{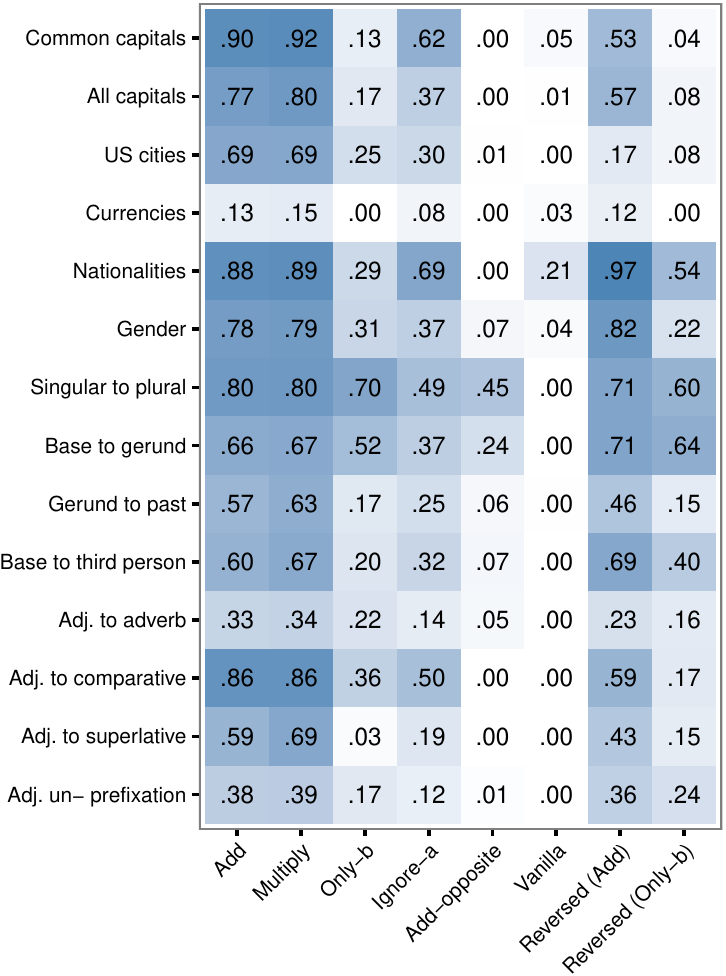}
    \caption{\label{fig:heatmap}Accuracy of all functions on space $s_5$.}
\end{figure}

Figure \ref{fig:heatmap} shows the success of all of the analogy functions in
recovering the intended analogy target $b^*$ in space $s_5$. In line with
\newcite{levy2014linguistic}, there was a slight advantage for
\textsc{Multiply} over \textsc{Add} (mean difference in accuracy: .03), as well
as dramatic variability across categories (ranging from $.13$ to $.90$ in \textsc{Add}).
This variability cuts across the distinction between the world-knowledge and
morphological categories; performance on currencies and adjectives-to-adverbs
was poor, while performance on capitals and comparatives was high.

Although \textsc{Add} and \textsc{Multiply} always outperformed the baselines,
the margin varied widely across categories. The most striking case is the
plurals category, where the accuracy of \textsc{Only-b} reached $.70$, and even
\textsc{Add-opposite} achieved a decent accuracy ($.45$). Taking $a^*$ but not
$a$ into account (\textsc{Ignore-a}) outperformed \textsc{Only-b} in ten out of
14 categories. Finally, the poor performance of \textsc{Vanilla} confirms that
   $a$, $a^*$ and $b$ must be excluded from the pool of potential answers for
   the offset method to work.  When these words were not excluded, the nearest
   neighbor of $a^* - a + b$ was $b$ in 93\% of the cases and $a^*$ in 5\% of
   the cases (it was never $a$).

\paragraph{Reversed analogies:}

Accuracy decreased in most categories when the direction of the analogy was
reversed (mean difference $-0.11$). The changes in the accuracy of \textsc{Add}
between the original and reversed problems were correlated across categories
with the changes in the performance of the \textsc{Only-b} baseline before and
after reversal (Pearson's $r = .72$). The fact that the performance of the
baseline that ignores the offset was a reliable predictor of the performance of
the offset method again suggests that the offset method when applied to the
\newcite{mikolov2013efficient} sets jointly evaluates the consistency of the
offsets and the probability that $b^*$ is the nearest neighbor of $b$.

The most dramatic decrease was in the US cities category (.69 to .17). This is
plausibly due to the fact that the city-to-state relation is a many-to-one
mapping; as such, the offsets derived from two specific city-states pairs
--- e.g., \textit{Sacramento:California} and \textit{Chicago:Illinois} --- are
unlikely to be exactly the same. Another sharp decrease was observed in the common
capitals category (.9 to .53), even though that category is presumably 
a one-to-one mapping.

\paragraph{Comparison across spaces:}

\begin{table}
    \scalebox{0.8}{
    \begin{tabular}{llll}
        \toprule
        Space & \textsc{Add} & \textsc{Add} - \textsc{Ignore-a} & \textsc{Add} - \textsc{Only-b} \\
        \midrule
        $s_2$ & .53 & .41 & .42 \\
        $s_5$ & .6 & .29 & .36 \\
        $s_{10}$ & .58 & .26 & .33 \\
        \bottomrule
    \end{tabular}
    }
    \caption{\label{table:overall}Overall scores and the advantage of \textsc{Add}
    over two of the baselines across spaces.}
\end{table}

\begin{figure}
    \centering
        \includegraphics[width=3in]{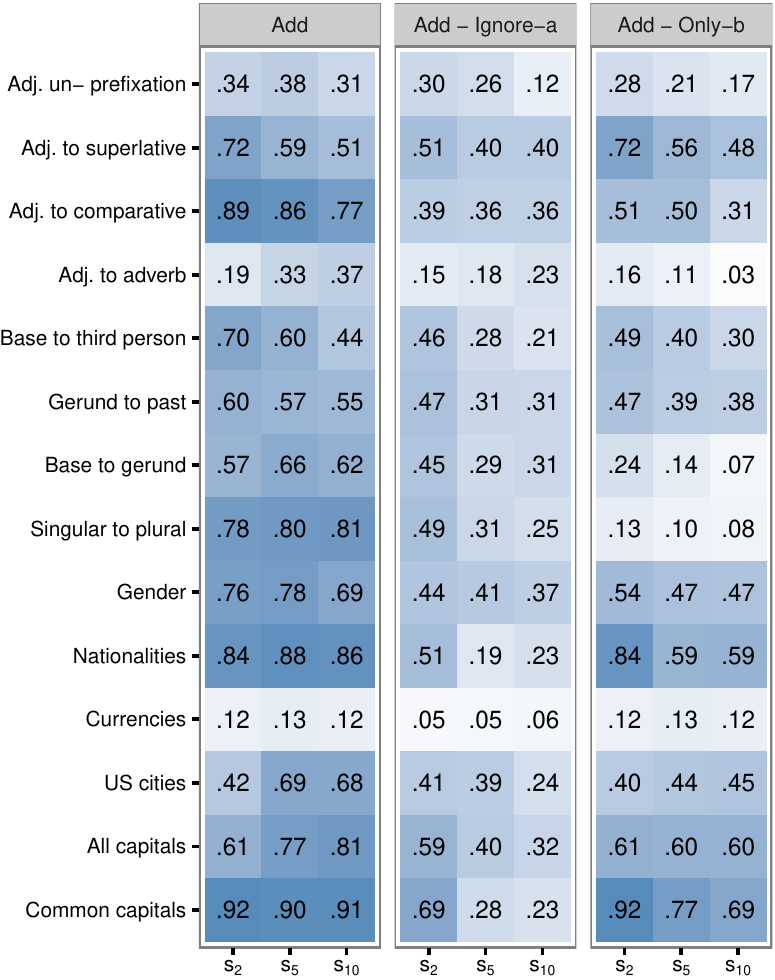}
    \caption{\label{fig:spaces}Comparison across spaces. The leftmost panel
    shows the accuracy of \textsc{Add}, and the next two panels show the
    improvement in accuracy of \textsc{Add} over the baselines.}
\end{figure}

The overall accuracy of \textsc{Add} was similar across spaces, with a small
advantage for $s_5$ (Table \ref{table:overall}). Yet the breakdown of the
results by category (Figure  \ref{fig:spaces}) shows that the similarity in
average performance across the spaces obscures differences across categories:
$s_2$ performed much better than $s_{10}$ in some of the morphological
inflection categories (e.g., .7 compared to .44 for the base-to-third-person
relation), whereas $s_{10}$ had a large advantage in some of the
world-knowledge categories (e.g., .68 compared to .42 in the US cities
category). The advantage of smaller window sizes in capturing ``syntactic''
information is consistent with previous studies
\cite{redington1998distributional,sahlgren2006word}. Note also that overall
accuracy figures are potentially misleading in light of the considerable
variability in the number of analogies in each category (see Table
\ref{table:analogies}): the ``all capitals'' category has a much greater effect
on overall accuracy than gender, for example.

Spaces also differed in how much \textsc{Add} improved over the baselines.  The
overall advantage over the baselines was highest for $s_2$ and lowest for
$s_{10}$ (Table \ref{table:overall}). In particular, although accuracy was
similar across spaces in the nationalities and common capitals categories, much
more of this accuracy was already captured by the \textsc{Ignore-a} baseline in
$s_{10}$ than in $s_2$ (Figure \ref{fig:spaces})

\section{Discussion}

The success of the offset method in solving word analogy problems has been
taken to indicate that systematic relations between words are represented in
the space as consistent vector offsets \cite{mikolov2013linguistic}. The
present note has examined potential difficulties with this interpretation.
A literal (``vanilla'') implementation of the method failed to perform the
task: the nearest neighbor of $a^* - a + b$ was almost always $b$ or
$a^*$.\footnote{A human with any reasonable understanding of the analogy task
is likely to also exclude $a$, $a^*$ and $b$ as possible responses, of course.
However, such heuristics that are baked into an analogy
solver, while likely to improve its performance, call into question the
interpretation of the success of the analogy solver as evidence for the
geometric organization of the underlying semantic space.} Even when those
candidates were excluded, some of the success of the method on the analogy sets
that we considered could also be obtained by baselines that ignored $a$ or even
both $a$ and $a^*$.  Finally, reversing the direction of the analogy affected
accuracy substantially, even though the same offset was involved in both
directions. 

The performance of the baselines varied widely across analogy categories.
Baseline performance was poor in the adjective-to-superlative relation, and was
very high in the plurals category (even when both $a$ and $a^*$ were ignored).
This suggests that analogy problems in the plural category category may not
measure whether the space encodes the single-to-plural relation as a vector
offset, but rather whether the plural form of a noun tends to be close in the
vector space to its singular form. Baseline performance varied across spaces as
well; in fact, the space with the weakest overall performance ($s_2$) showed
the largest increases over the baselines, and therefore the most evidence for
consistent offsets.

We suggest that future studies employing the analogy task report the
performance of the simple baselines we have suggested, in particular
\textsc{Only-b} and possibly also \textsc{Ignore-a}. Other methods for
evaluating the consistency of vector offsets may be less vulnerable to trivial
responses and neighborhood structure, and should be considered instead of the
offset method \cite{dunbar2015quantitative}.

Our results also highlight the difficulty in comparing spaces based on accuracy
measures averaged across heterogeneous and unbalanced analogy sets
\cite{gladkova2016analogy}. Spaces with similar overall accuracy can vary in
their success on particular categories of analogies; effective representations
of ``world-knowledge'' information are likely to be useful for different
downstream tasks than effective representations of formal linguistic
properties. Greater attention to the fine-grained strengths of particular
spaces may lead to the development of new spaces that combine these strengths.

\section*{Acknowledgments}

I thank Ewan Dunbar, Emmanuel Dupoux, Omer Levy and Benjamin Spector for
comments and discussion. This research was supported by the European Research
Council (grant ERC-2011-AdG 295810 BOOTPHON) and the Agence Nationale pour la
Recherche (grants ANR-10-IDEX-0001-02 PSL and ANR-10-LABX-0087 IEC).

\bibliographystyle{acl2016}
%\bibliography{../../../../cogneurolang.bib}
\bibliography{bibexport}
\end{document}